\documentclass[11pt]{article}
\usepackage{coling2018}
\usepackage{times}
\usepackage{url}
\usepackage{latexsym}
\usepackage{amsmath}
\usepackage{multirow}
\usepackage{color}
\usepackage{subfigure}
\usepackage{url}
\usepackage{CJK}
\usepackage{amsfonts}
\usepackage{graphicx} 
\usepackage{url}

\title{Subword-augmented Embedding for Cloze Reading Comprehension}

\author{Zhuosheng Zhang$^{1,2,}$\thanks{$\ $ These authors contribute equally. $\dagger$ Corresponding author. This paper was partially supported by
		National Key Research and Development Program of China (No. 2017YFB0304100),
		National Natural Science Foundation of China (No. 61672343 and No. 61733011),
		Key Project of National Society Science Foundation of China (No. 15-ZDA041),
		The Art and Science Interdisciplinary Funds of Shanghai Jiao Tong University (No. 14JCRZ04).} , Yafang Huang$^{1,2,*}$, Hai Zhao$^{1,2,\dagger}$\\
	$^{1}$Department of Computer Science and Engineering, Shanghai Jiao Tong University \\
	$^{2}$Key Laboratory of Shanghai Education Commission for Intelligent Interaction \\ and Cognitive Engineering, Shanghai Jiao Tong University, Shanghai, 200240, China\\
	{\tt \{zhangzs, huangyafang\}@sjtu.edu.cn, zhaohai@cs.sjtu.edu.cn}
}

\date{}

\begin{document}
\maketitle
\begin{abstract}
	Representation learning is the foundation of machine reading comprehension. In state-of-the-art models, deep learning methods broadly use word and character level representations. However, character is not naturally the minimal linguistic unit. In addition, with a simple concatenation of character and word embedding, previous models actually give suboptimal solution. In this paper, we propose to use subword rather than character for word embedding enhancement. We also empirically explore different augmentation strategies on  \emph{subword-augmented embedding} to enhance the cloze-style reading comprehension model (reader). In detail, we present a reader that uses subword-level representation to augment word embedding with a short list to handle rare words effectively. A thorough examination is conducted to evaluate the comprehensive performance and generalization ability of the proposed reader. Experimental results show that the proposed approach helps the reader significantly outperform the state-of-the-art baselines on various public datasets.
\end{abstract}

\section{Introduction}
\blfootnote{
	%
	%
	%
	%
	%
	%
	This work is licensed under a Creative Commons 
	Attribution 4.0 International License.
	License details:
	\url{http://creativecommons.org/licenses/by/4.0/}
}

\noindent A recent hot challenge is to train machines to read and comprehend human languages. Towards this end, various machine reading comprehension datasets have been released, including cloze-style \cite{hermann2015teaching,hill2015goldilocks,Cui2016Consensus} and user-query types \cite{Joshi2017TriviaQA,rajpurkar2016squad}. Meanwhile, a number of deep learning models are designed to take up the challenges, most of which focus on attention mechanism \cite{Wang2017Gated,Seo2016Bidirectional,Cui2016Attention,kadlec2016text,Dhingra2017Gated,zhang2018OneShot}. However, how to represent word in an effective way remains an open problem for diverse natural language processing tasks, including machine reading comprehension for different languages. Particularly, for a language like Chinese with a large set of characters (typically, thousands of), lots of which are semantically ambiguous, using either word-level or character-level embedding alone to build the word representations would not be accurate enough. This work especially focuses on a cloze-style reading comprehension task over fairy stories, which is highly challenging due to diverse semantic patterns with personified expressions and reference. 

In real practice, a reading comprehension model or system which is often called \emph{reader} in literatures easily suffers from out-of-vocabulary (OOV) word issues, especially for the cloze-style reading comprehension tasks when the ground-truth answers tend to include rare words or named entities (NE), which are hardly fully recorded in the vocabulary. This is more challenging in Chinese. There are over 13,000 characters in Chinese\footnote{Refer to the statistics of Xinhua Dictionary, version 11, published by The Commercial Press in 2014.} while there are only 26 letters in English without regard to punctuation marks. If a reading comprehension system cannot effectively manage the OOV issues, the performance will not be semantically accurate for the task. 

Commonly, words are represented as vectors using either word embedding or character embedding. For the former, each word is mapped into low dimensional dense vectors from a lookup table.  Character representations are usually obtained by applying neural networks on the character sequence of the word, and their hidden states are obtained to form the representation. Intuitively, word-level representation is good at catching global context and dependency relationships between words, while character embedding helps for dealing with rare word representation.

However, the minimal meaningful unit below word usually is not character, which motivates researchers to explore the potential unit (subword) between character and word to model sub-word morphologies or lexical semantics. In fact, morphological compounding (e.g. \emph{sunshine} or \emph{playground}) is one of the most common and productive methods of word formation across human languages, which inspires us to represent word by meaningful sub-word units. Recently, researchers have started to work on morphologically informed word embeddings \cite{Botha2014Compositional,Cao2016A}, aiming at better capturing syntactic, lexical and morphological information. With ready subwords, we do not have to work with characters, and segmentation could be stopped at the subword-level to reach a meaningful representation. 

In this paper, we present various simple yet accurate subword-augmented embedding (SAW) strategies and propose SAW Reader as an instance. Specifically, we adopt subword information to enrich word embedding and survey different SAW operations to integrate word-level and subword-level embedding for a fine-grained representation. To ensure adequate training of OOV and low-frequency words, we employ a short list mechanism. Our evaluation will be performed on three public Chinese reading comprehension datasets and one English benchmark dataset for showing our method is also effective in multi-lingual case. 

\begin{figure*}
	\centering
	\includegraphics[width=1.0\textwidth]{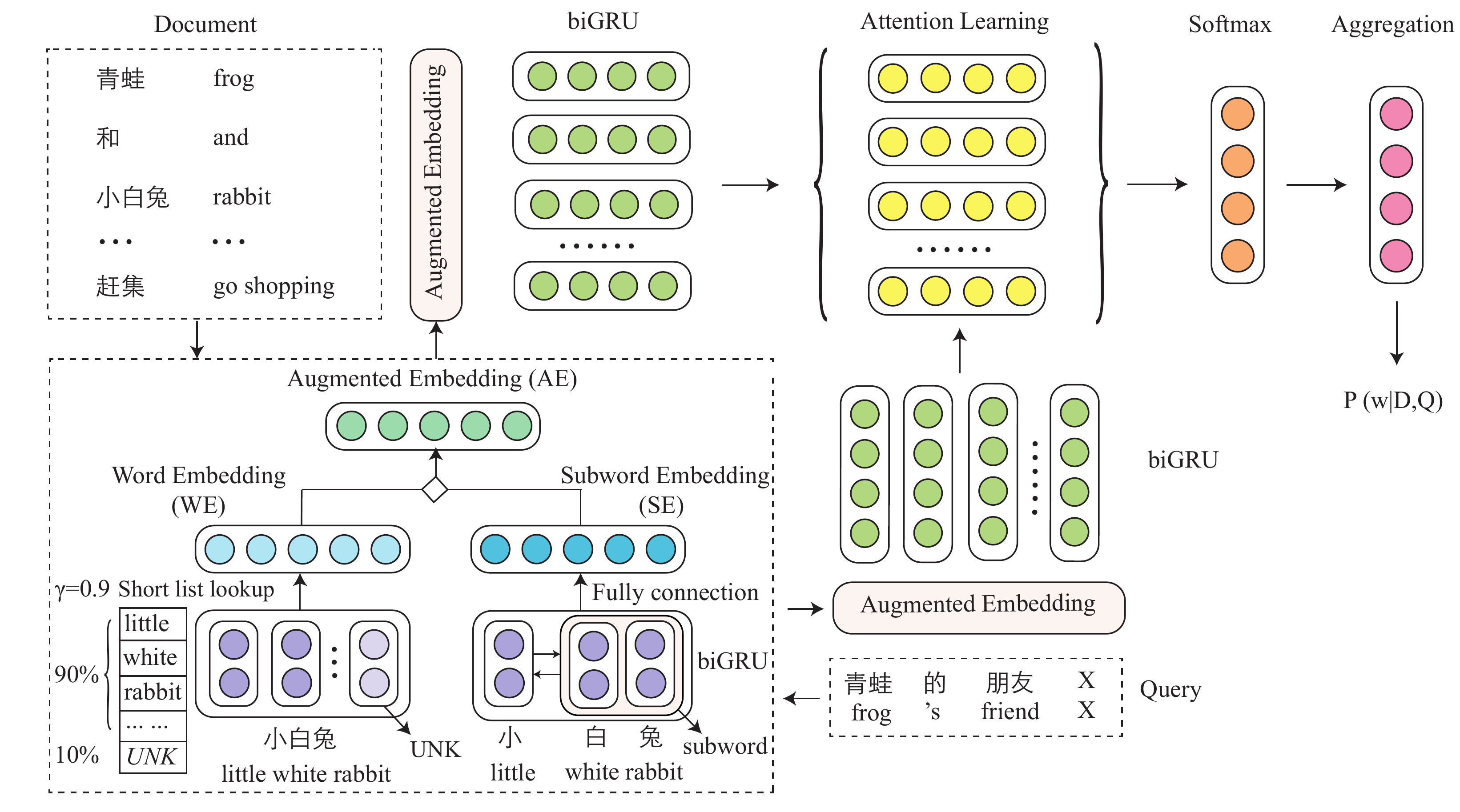}
	\caption{Architecture of the proposed Subword-augmented Embedding Reader (SAW Reader).}
	\label{fig:framework}
\end{figure*}

\section{The Subword-augmented Word Embedding}
The concerned reading comprehension task can be roughly categorized as user-query type and cloze-style according to the answer form. Answers in the former are usually a span of texts while in the cloze-style task, the answers are words or phrases which lets the latter be the harder-hit area of OOV issues, inspiring us to select the cloze-style as our testbed for SAW strategies. Our preliminary study shows even for the advanced word-character based GA reader, OOV answers still account for nearly 1/5 in the error results. This also motivates us to explore better representations to further performance improvement.

The cloze-style task in this work can be described as a triple $<D, Q, A>$, where $D$ is a document (context), $Q$ is a query over the contents of $D$, in which a word or phrase is the right answer $A$.
This section will introduce the proposed SAW Reader in the context of cloze-style reading comprehension. Given the triple $<D, Q, A>$, the SAW Reader will be built in the following steps.

\subsection{BPE Subword Segmentation}
Word in most languages usually can be split into meaningful subword units despite of the writing form. For example, ``\emph{indispensable}" could be split into the following subwords: $<in, disp, ens, able>$. 

In our implementation, we adopt Byte Pair Encoding (BPE) \cite{Gage1994A} which is a simple data compression technique that iteratively replaces the most frequent pair of bytes in a sequence by a single, unused byte. 
BPE allows for the representation of an open vocabulary through a fixed-size vocabulary of variable-length character sequences, making it a very suitable word segmentation strategy for neural network models. 

The generalized framework can be described as follows. Firstly, all the input sequences (strings) are tokenized into a sequence of single-character subwords, then we repeat,
\begin{enumerate}
	\item Count all bigrams under the current segmentation status of all sequences.
	\item Find the bigram with the highest frequency and merge them in all the sequences. Note the segmentation status is updating now.
	\item If the merging times do not reach the specified number, go back to 1, otherwise the algorithm ends.
\end{enumerate}	

In \cite{Sennrich2015Neural}, BPE is adopted to segment infrequent words into sub-word units for machine translation. However, there is a key difference between the motivations for subword segmentation. We aim to refine the word representations by using subwords, for both frequent and infrequent words, which is more generally motivated. To this end, we adaptively tokenize words in multi-granularity by controlling the merging times. 

\subsection{Subword-augmented Word Embedding}
Our subwords are also formed as character n-grams, do not cross word boundaries. After using unsupervised segmentation methods to split each word into a subword sequence, an augmented embedding (AE) is to straightforwardly integrate word embedding $WE(w)$ and subword embedding $SE(w)$ for a given word $w$.
\begin{align*}	
AE(w)= WE(w) \diamond  SE(w) 
\end{align*}
where $\diamond$ denotes the detailed integration operation. In this work, we investigate concatenation (\emph{concat}), element-wise summation (\emph{sum}) and element-wise multiplication (\emph{mul}). Thus, each document $D$ and query $Q$ is represented as $\mathbb{R}^{d \times k}$ matrix where $d$ denotes the dimension of word embedding and $k$ is the number of words in the input. 

Subword embedding could be useful to refine the word embedding in a finer-grained way, we also consider improving word representation from itself. For quite a lot of words, especially those rare ones, their word embedding is extremely hard to learn due to the data sparse issue. Actually, if all the words in the dataset are used to build the vocabulary, the OOV words from the test set will not obtain adequate training. If they are initiated inappropriately, either with relatively high or low weights, they will harm the answer prediction. To alleviate the OOV issues, we keep a short list $H$ for specific words.
\begin{align*}	
H = \{w_{1}, w_{2}, \dots, w_{n}\}
\end{align*}

If $w$ is in $H$, the immediate word embedding $WE(w)$ is indexed from word lookup table $M^{w} \in \mathbb{R}^{d \times s}$ where $s$ denotes the size (recorded words) of lookup table. Otherwise, it will be represented as the randomly initialized default word (denoted by a specific mark $UNK$). Note that, this is intuitively like ``guessing'' the possible unknown words (which will appear during test) from the vocabulary during training and only the word embedding of the OOV words will be replaced by $UNK$ while their subword embedding $SE(w)$ will still be processed using the original word. In this way, the OOV words could be tuned sufficiently with expressive meaning after training. During test, the word embedding of unknown words would not severely bias its final representation. Thus, $AE$($w$) can be rewritten as
\begin{align*}	
AE(w)=\left\{
\begin{array}{ll}
WE(w) \diamond  SE(w)   &   \text{if} \ w \in H\\
UNK \diamond  SE(w) & \text{otherwise}\\
\end{array} \right.
\end{align*}

In our experiments, the short list is determined according to the word frequency. Concretely, we sort the vocabulary according to the word frequency from high to low. A frequency filter ratio $\gamma$ is set to filter out the low-frequency words (rare words) from the lookup table. For example, $\gamma$=0.9 means the least frequent 10\% words are replaced with the default UNK notation.

The subword embedding $SE(w)$ is generated by taking the final outputs of a bidirectional gated recurrent unit (GRU) \cite{Cho2014Learning} applied to the embeddings from a lookup table of subwords. The structure of GRU used in this paper are described as follows.
\begin{align*}
r_{t} &=\sigma (W_{r}x_{t}+U_{r}h_{t-1}+b_{r}), \\
z_{t} &=\sigma (W_{z}x_{t}+U_{z}h_{t-1}+b_{z}), \\
\tilde{h}_{t} &=\textup{tanh}(W_{h}x_{t}+U_{h}(r{t}\odot h_{t-1})+b_{h}) \\
h_{t} & = (1-z_{t})\odot h_{t-1}+z_{t}\odot \tilde{h}_{t}
\end{align*}
where $\odot$ denotes the element-wise multiplication. $r_{t}$ and $z_{t}$ are the reset and update gates respectively, and $\tilde{h}_{t}$ are the hidden states. A bi-directional GRU (BiGRU) processes the sequence in both forward and backward directions. Subwords of each word are successively fed to forward GRU and backward GRU to obtain the internal features of two directions. The output for each input is the concatenation of the two vectors from both directions: $\overleftrightarrow{h_{t}} = \overrightarrow{h_{t}}\parallel \overleftarrow{h_{t}}$. Then, the output of BiGRUs is passed to a fully connected layer to obtain the final subword embedding $SE(w)$. 
\begin{align*}
SE(w) = W \overleftrightarrow{h_{t}} + b
\end{align*}

\begin{table*}
	\centering
	{
		\begin{tabular}{cccccccc}
			\hline
			\hline
			& \multicolumn{3}{c}{CMRC-2017} & \multicolumn{3}{c}{PD}  & CFT  \\
			&Train & Valid & Test & Train & Valid & Test & human  \\
			\hline
			\# Query & 354,295 & 2,000 & 3,000 &870,710 & 3,000 & 3,000  & 1,953\\
			Max \# words in docs  & 486 & 481  & 484 & 618 & 536 & 634 &  414\\
			Max \# words in query & 184 & 72  & 106 & 502 & 153 & 265 &   92\\
			Avg \# words in docs & 324 & 321  & 307 & 379 & 425 & 410 &   153\\
			Avg \# words in query & 27 & 19 & 23 & 38 & 38 & 41 & 20\\
			\# Vocabulary  & 94,352 & 21,821 & 38,704 & 248,160 & 536 & 634 & 414\\
			\hline
			\hline
		\end{tabular}
	}
	\caption{\label{tab:dataset} Data statistics of CMRC-2017, PD and CFT.}
\end{table*}

\subsection{Attention Module}
Our attention module is based on the Gated attention Reader (GA Reader) proposed by \cite{Dhingra2017Gated}. We choose this model due to its simplicity with comparable performance so that we can focus on the effectiveness of SAW strategies. This module can be described in the following two steps.
After augmented embedding, we use two BiGRUs to get contextual representations of the document and query respectively, where the representation of each word is formed by concatenating the forward and backward hidden states. 
\begin{align*}
H_{q} &= \textup{BiGRU}(Q) \\
H_{d} &= \textup{BiGRU}(D)
\end{align*}
For each word $d_{i}$ in $H_{d}$, we form a word-specific representation of the query $q_{i} \in H_{q}$ using soft attention, and then adopt element-wise product to multiply the query representation with the document word representation.
\begin{align*}
\alpha_{i} &= softmax(H_{q}^{\top}d_{i}) \\
\beta_{i} &= Q\alpha_{i} \\
x_{i} &= d_{i}\odot \beta_{i} \\
\end{align*}
where $\odot$ denotes the multiplication operator to model the interactions between $d_{i}$ and $q_{i}$. Then, the document contextual representation ${\tilde H}_{d} = \{x_{1}, x_{2},\dots, x_{k}\}$ is gated by query representation.

Suppose the network has $K$ layers. At each layer, the document representation ${\tilde H}_{d}$ is updated through above attention learning. After going through all the layers, our model comes to answer prediction phase. We use all the words in the document to form the candidate set $C$. Let $q_{t}$ denote the $t$-th intermediate output of query representation $H_{q}$ and $H_{D}$ represent the full output of document representation $\tilde{H}_{d}$. The probability of each candidate word $w \in C$ as being the answer is predicted using a softmax layer over the inner-product between $q_{t}$ and $H_{D}$.
\begin{align*}
p = softmax ((q_t)^{\top} H_{D})
\end{align*}
where vector $p$ denotes the probability distribution over all the words in the document.
Note that each word may occur several times in the document.
Thus, the probabilities of each candidate word occurring in different positions of the document are summed up for final prediction.
\begin{align*}
P(w|D,Q) \propto \sum_{i \in I(w,D)}p_{i}
\end{align*}
where $I(w,d)$ denotes the set of positions that a particular word $w$ occurs in the document $D$. The training objective is to maximize $\log P(A|D,Q)$ where $A$ is the correct answer. 

Finally, the candidate word with the highest probability will be chosen as the predicted answer.
\begin{align*}
A^{\ast }=\textup{argmax}_{w \in C} P(w|D,Q)
\end{align*}

Different from recent work employing complex attention mechanisms \cite{Wang2017Gated,Cui2016Attention,sordoni2016iterative}, our attention mechanism is much more simple with comparable performance so that we can focus on the effectiveness of SAW strategies.

\section{Experiments}

\subsection{Dataset and Settings}
To verify the effectiveness of our proposed model, we conduct multiple experiments on three Chinese Machine Reading Comprehension datasets, namely CMRC-2017 \cite{Cui2017Dataset}, People's Daily (PD) and Children Fairy Tales (CFT) \cite{Cui2016Consensus}\footnote{Note that the test set of CMRC-2017 and human evaluation test set (Test-human) of CFT are harder for the machine to answer because the questions are further processed manually and may not be accordance with the pattern of automatic questions.}. In these datasets, a story containing consecutive sentences is formed as the \emph{Document} and one of the sentences is either automatically or manually selected as the \emph{Query} where one token is replaced by a placeholder to indicate the \emph{answer} to fill in. Table \ref{tab:dataset} gives data statistics. Different from the current cloze-style datasets for English reading comprehension, such as CBT, Daily Mail and CNN \cite{hermann2015teaching}, the three Chinese datasets do not provide candidate answers. Thus, the model has to find the correct answer from the entire document.

Besides, we also use the Children's Book Test (CBT) dataset \cite{hill2015goldilocks} to test the generalization ability in multi-lingual case. We only focus on subsets where the answer is either a common noun (CN) or NE which is more challenging since the answer is likely to be rare words. We evaluate all the models in terms of accuracy, which is the standard evaluation metric for this task.  

Throughout this paper, we use the same model setting to make fair comparisons. According to our preliminary experiments, we report the results based on the following settings. The default integration strategy is \emph{element-wise product}. Word embeddings were 200$d$ and pre-trained by word2vec \cite{mikolov:2013} toolkit on \emph{Wikipedia} corpus\footnote{\url{https://dumps.wikimedia.org/} }. Subword embedding were 100$d$ and randomly initialized with the uniformed distribution in the interval [-0:05; 0:05]. Our model was implemented using the Theano\footnote{\url{https://github.com/Theano/Theano}} and Lasagne Python libraries\footnote{\url{https://github.com/Lasagne/Lasagne}}. We used stochastic gradient descent with ADAM updates for optimization \cite{kingma2014adam}. The batch size was 64 and the initial learning rate was 0.001 which was halved every epoch after the second epoch. We also used gradient clipping with a threshold of 10 to stabilize GRU training \cite{Pascanu2013On}. We use three attention layers for all experiments. The GRU hidden units for both the word and subword representation were 128. The default frequency filter proportion was 0.9 and the default merging times of BPE was 1,000. We also apply dropout between layers with a dropout rate of 0.5 \footnote{Our code is available at: \url{https://github.com/cooelf/subMrc}}.

\subsection{Main Results}

\begin{table}
	\centering
	{
		\begin{tabular}{l|c|c}
			\hline
			\hline
			\multirow{2}{*}{Model}  & \multicolumn{2}{c}{CMRC-2017} \\
			& Valid & Test \\
			\hline
			Random Guess \dag & 1.65 & 1.67\\
			Top Frequency \dag  & 14.85 & 14.07  \\
			AS Reader \dag & 69.75 & 71.23 \\
			GA Reader &  72.90 & 74.10 \\
			\cline{1-3}
			SJTU BCMI-NLP \dag & 76.15 & 77.73 \\
			6ESTATES PTE LTD \dag & 75.85 & 74.73  \\
			Xinktech \dag & 77.15 & 77.53 \\
			Ludong University \dag & 74.75 & 75.07 \\
			ECNU \dag & 77.95 & 77.40 \\
			WHU \dag & 78.20 & 76.53 \\
			\cline{1-3}
			SAW Reader & \textbf{78.95}  & \textbf{78.80} \\
			\hline
			\hline
		\end{tabular}
	}
	\caption{\label{tab:cmrc} Accuracy on CMRC-2017 dataset. Results marked with $\dag$ are from the latest official CMRC-2017 Leaderboard \footnotemark[7]. The best results are in bold face.}
\end{table}
\footnotetext[7]{\url{http://www.hfl-tek.com/cmrc2017/leaderboard.html}}

\paragraph{CMRC-2017} Table \ref{tab:cmrc} shows our results on CMRC-2017 dataset, which shows that our SAW Reader (mul) outperforms all other single models on the test set, with 7.57\% improvements compared with Attention Sum Reader (AS Reader) baseline. Although WHU's model achieves the best besides our model on the valid set with only 0.75\% below ours, their result on the test set is lower than ours by 2.27\%, indicating our model has a satisfactory generalization ability.

\begin{table}
	\centering
	{
		\begin{tabular}{l|c|c|c}
			\hline
			\hline
			\multirow{2}{*}{Model} & \multirow{2}{*}{Operation}  & \multicolumn{2}{c}{CMRC-2017} \\
			&	& Valid & Test \\
			\hline
			& concat  & 74.80 & 75.13 \\
			Word + Char & sum  & 75.40  & 75.53  \\
			& mul  &  77.80  & 77.93   \\
			\hline
			& concat  &  75.95 &  76.43 \\
			Word + BPE & sum   & 76.20   &  75.83  \\
			& mul  & \textbf{78.95} & \textbf{78.80} \\
			\hline
			\hline
		\end{tabular}
	}
	\caption{\label{tab:oper} Case study on CMRC-2017.}
\end{table}

We also list different integration operations for word and subword embeddings. Table \ref{tab:oper} shows the comparisons. From the results, we can see that Word + BPE outperforms Word + Char which indicates subword embedding works essentially. We also observe that \emph{mul} outperforms the other two operations, \emph{concat} and \emph{sum}. This reveals that \emph{mul} might be more informative than \emph{concat} and \emph{sum} operations. The superiority might be due to element-wise product being capable of modeling the interactions and eliminating distribution differences between word and subword embedding. Intuitively, this is also similar to endow subword-aware ``attention'' over the word embedding. In contrast, concatenation operation may cause too high dimension, which leads to serious over-fitting issues, and sum operation is too simple to prevent from detailed information losing. 

\begin{table}
	\centering 
	{
		\begin{tabular}{l|c|c|c}
			\hline
			\hline
			\multirow{2}{*}{Model}  & \multicolumn{2}{c}{PD}  & CFT  \\
			& Valid & Test &Test-human  \\
			
			\hline
			AS Reader &  64.1 & 67.2 & 33.1  \\
			GA Reader &  67.2 & 69.0 & 36.9  \\
			CAS Reader &  65.2 & 68.1 & 35.0  \\
			\hline
			SAW Reader & \textbf{72.8} & \textbf{75.1} & \textbf{43.8}    \\
			\hline
			\hline
		\end{tabular}
	}
	\caption{\label{tab:pdcftresult} Accuracy on PD and CFT datasets. Results of AS Reader and CAS Reader are from \cite{Cui2016Consensus}.}
\end{table}

\paragraph{PD \& CFT} Since there is no training set for CFT dataset, our model is trained on PD training set. Note that the CFT dataset is harder for the machine to answer because the test set is further processed by human evaluation, and may not be accordance with the pattern of PD dataset. The results on PD and CFT datasets are listed in Table \ref{tab:pdcftresult}. As we see that, our SAW Reader significantly outperforms the CAS Reader in all types of testing, with improvements of 7.0\% on PD and 8.8\% on CFT test sets, respectively. Although the domain and topic of PD and CFT datasets are quite different, the results indicate that our model also works effectively for out-of-domain learning.

\begin{table}
	\centering 
	{
		\begin{tabular}{l|c|c|c|c}
			\hline
			\hline
			\multirow{2}{*}{Model}  & \multicolumn{2}{c}{CBT-NE}  & \multicolumn{2}{c}{CBT-CN} \\
			& Valid & Test   & Valid & Test  \\
			\hline
			Human \ddag &- & 81.6 & - & 81.6 \\
			\hline
			LSTMs \ddag & 51.2 & 41.8 & 62.6 & 56.0 \\ 
			MemNets \ddag & 70.4 & 66.6 & 64.2 & 63.0 \\
			AS Reader \ddag & 73.8 & 68.6 & 68.8 & 63.4 \\
			Iterative Attentive Reader  \ddag & 75.2 & 68.2 & 72.1 & 69.2 \\
			EpiReader \ddag & 75.3 & 69.7 & 71.5 & 67.4 \\
			AoA Reader \ddag & 77.8 & 72.0 & 72.2 & 69.4 \\
			NSE \ddag & 78.2 & 73.2 & 74.3 & 71.9 \\
			FG Reader \ddag & \textbf{79.1} & \textbf{75.0} & \textbf{75.3}&  \textbf{72.0} \\
			GA Reader \ddag &76.8 & 72.5  & 73.1 & 69.6 \\
			\hline
			SAW Reader& 78.5 & 74.9 & 75.0 & 71.6\\
			\hline
			\hline
		\end{tabular}
	}
	\caption{\label{tab:cbt} Accuracy on CBT dataset. Results marked with $\ddag$ are of previously published works \cite{Dhingra2017Gated,Cui2016Consensus,Yang2016Words}.}
\end{table}	

\paragraph{CBT} To verify if our method can only work for Chinese, we also evaluate the effectiveness of the proposed method on benchmark English dataset. We use CBT dataset as our testbed to evaluate the performance. For a fair comparison, we simply set the same parameters as before. Table \ref{tab:cbt} shows the results. We observe that our model outperforms most of the previously public works, with 2.4 \% gains on the CBT-NE test set compared with GA Reader which adopts word and character embedding concatenation. Our SAW Reader also achieves comparable performance with FG Reader who adopts neural gates to combine word-level and character-level representations with assistance of extra features including NE, POS and word frequency while our model is much simpler and faster. This result shows our SAW Reader is not restricted to Chinese reading comprehension, but also for other languages.

\section{Analysis}

\subsection{Merging Times of BPE}
\begin{figure}[h]\small\centering
	\begin{minipage}[t]{0.42\linewidth}
		\centering
		\includegraphics[width=1\textwidth]{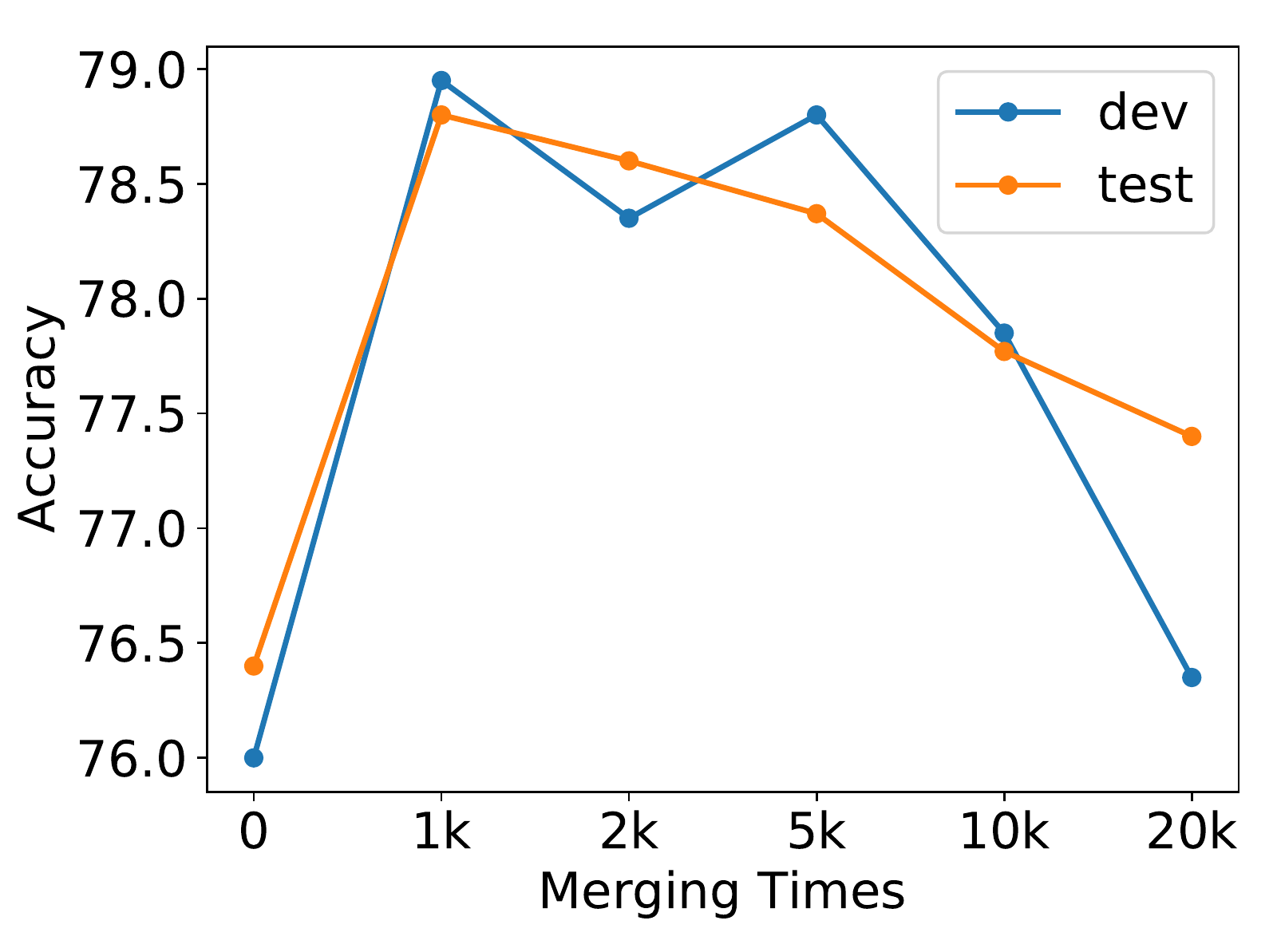}
		\caption{Case study of the subword vocabulary size of BPE.}
		\label{fig:bpe}
	\end{minipage} \quad
	\begin{minipage}[t]{0.42\linewidth}
		\centering
		\includegraphics[width=1\textwidth]{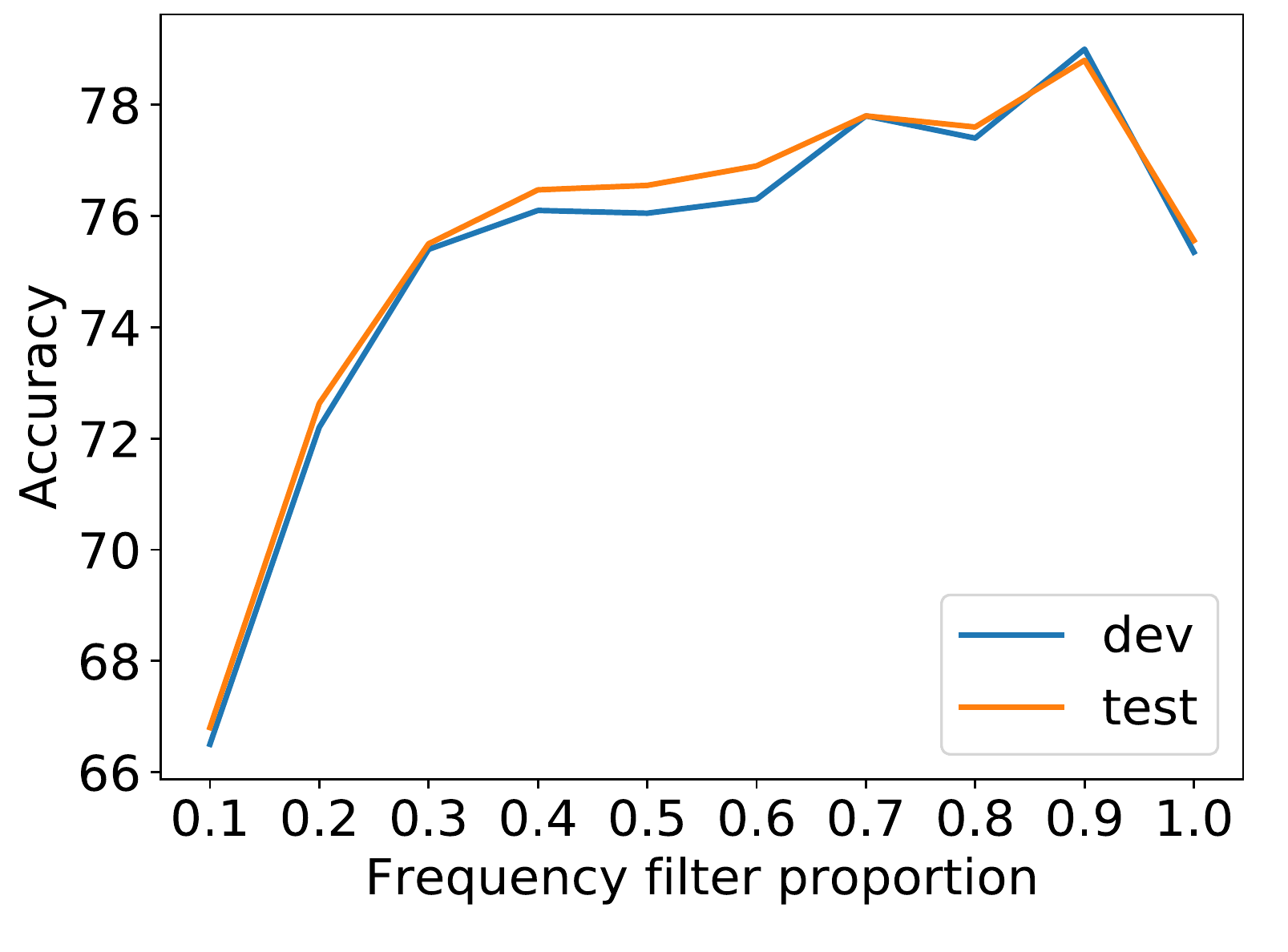}
		\caption{Quantitative study on the influence of the short list.}
		\label{fig:proportion}
	\end{minipage}
\end{figure}

The vocabulary size could seriously involve the segmentation granularity. For BPE segmentation, the resulted subword vocabulary size is equal to the merging times plus the number of single-character types. To have an insight of the influence, we adopt merge times  from 0 to 20$k$, and conduct quantitative study on CMRC-2017 for BPE segmentation. Figure \ref{fig:bpe} shows the results. We observe that when the vocabulary size is 1$k$, the models could obtain the best performance. The results indicate that for a task like reading comprehension the subwords, being a highly flexible grained representation between character and word, tends to be more like characters instead of words. However, when the subwords completely fall into characters, the model performs the worst. This indicates that the balance between word and character is quite critical and an appropriate grain of character-word segmentation could essentially improve the word representation.

\subsection{Filter Mechanism}
To investigate the impact of the short list to the model performance, we conduct quantitative study on the filter ratio from [0.1, 0.2, \dots, 1]. The results on the CMRC-2017 dataset are depicted in Figure \ref{fig:proportion}. As we can see that when $\gamma=0.9$ our SAW reader can obtain the best performance, showing that building the vocabulary among all the training set is not optimal and properly reducing the frequency filter ratio can boost the accuracy. This is partially attributed to training the model from the full vocabulary would cause serious over-fitting as the rare words representations can not obtain sufficient tuning. If the rare words are not initialized properly, they would also bias the whole word representations. Thus a model without OOV mechanism will fail to precisely represent those inevitable OOV words from test sets.

\subsection{Subword-Augmented Representations} 

In text understanding tasks, if the ground-truth answer is OOV word or contains OOV word(s), the performance of deep neural networks would severely drop due to the incomplete representation, especially for cloze-style reading comprehension task where the answer is only one word or phrase. In CMRC-2017, we observe questions with OOV answers (denoted as ``OOV questions") account for 17.22\% in the error results of the best Word + Char embedding based model. With BPE subword embedding, 12.17\% of these ``OOV questions" could be correctly answered. This shows the subword representations could be essentially useful for modeling rare and unseen words. 

\begin{figure*}\small\centering
	
	\subfigure[Embedding of Document and query]{
		\begin{minipage}[b]{0.455\textwidth}
			\includegraphics[width=1\textwidth]{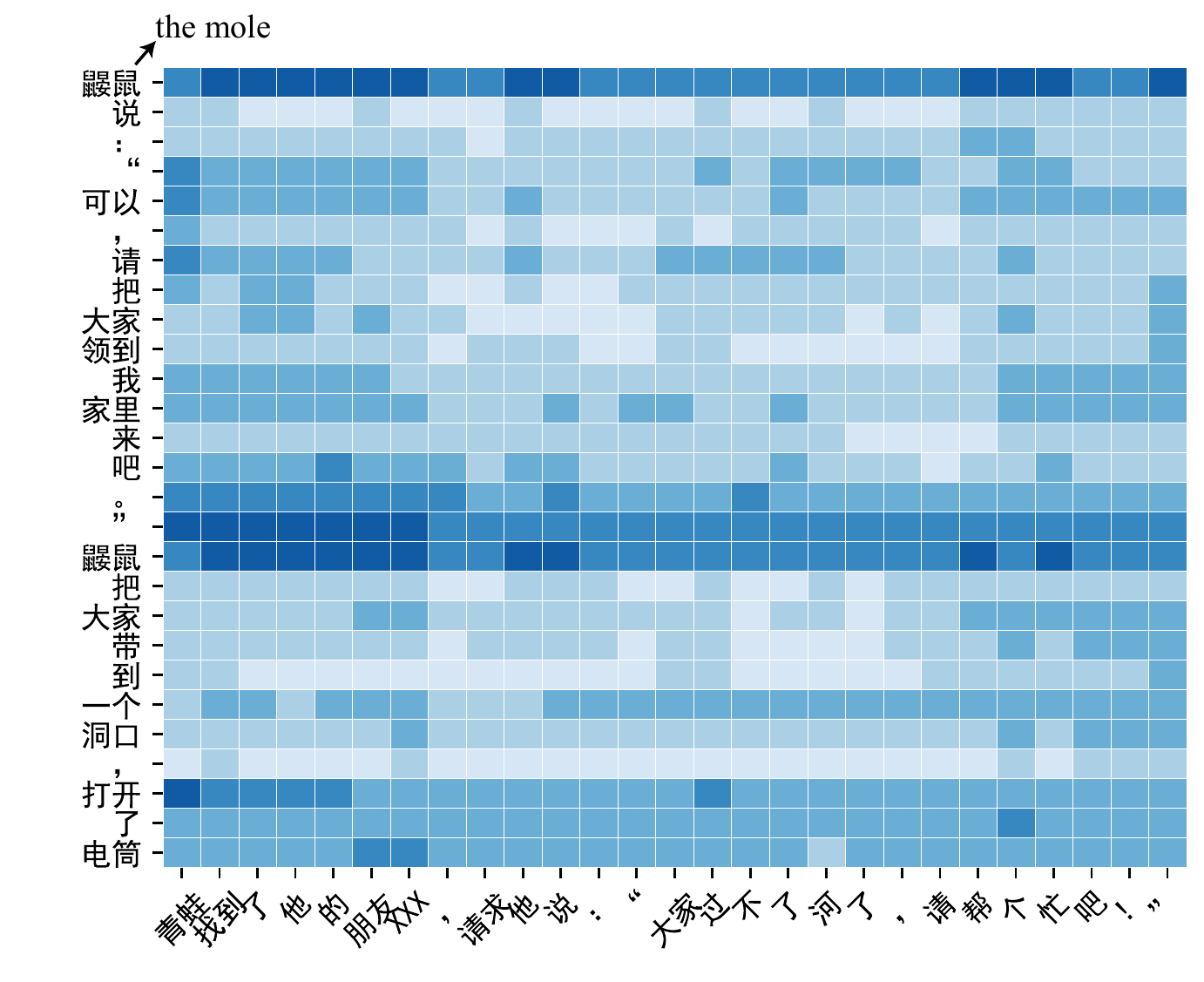}
		\end{minipage}
	}
	\subfigure[Final document and query representation]{
		\begin{minipage}[b]{0.5\textwidth}
			\includegraphics[width=1\textwidth]{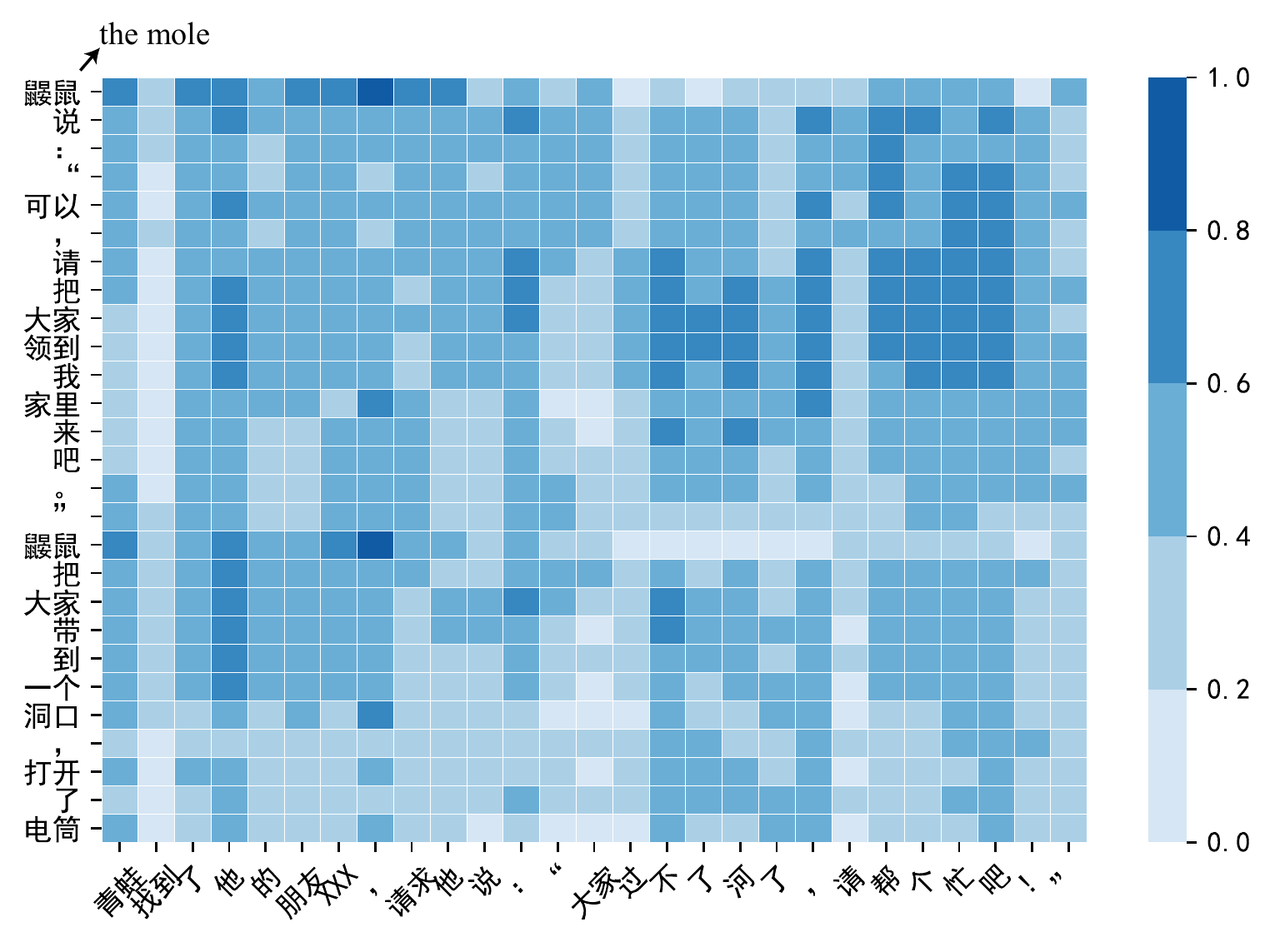}
		\end{minipage}
	}
	\\\begin{flushleft}
	\scriptsize{\emph{Doc (extract): The mole said, "That's fine, please bring them to my house." The mole took everyone to a hole, turned on the flashlight and asked the little white rabbit, the hedgehog, the big ant and the frog to follow him, saying, "Don't be afraid, just go ahead."
			\\Query: The frog found his friend \underline{\hbox to 8mm{}} and told him, We cannot get across the river. Please give us a hand!} }\end{flushleft}
	
	\caption{Pair-wise attention visualization. } \label{fig:attention}	
\end{figure*}

To analyze the reading process of SAW Reader, we draw the attention distributions at intermediate layers as shown in Figure \ref{fig:attention}. We observe the salient candidates in the document can be focused after the pair-wise matching of document and query and the right answer (``\emph{The mole}") could obtain a high weight at the very beginning. After attention learning, the key evidence of the answer would be collected and irrelevant parts would be ignored. This shows our SAW Reader is effective at selecting the vital points at the fundamental embedding layer, guiding the attention layers to collect more relevant pieces.

\section{Related Work}

\subsection{Machine Reading Comprehension}

Recently, many deep learning models have been proposed for reading comprehension \cite{sordoni2016iterative,Trischler2016Natural,Wang2016Machine,Munkhdalai2016Reasoning,Wang2017Conditional,Dhingra2017Gated,zhang2018DUA,Yizhong2018Multi}. Notably, \newcite{Chen2016A} conducted an in-depth and thoughtful examination on the comprehension task based on an attentive neural network and an entity-centric classifier with a careful analysis based on handful features. \newcite{kadlec2016text} proposed the Attention Sum Reader (AS Reader) that uses attention to directly pick the answer from the context, which is motivated by the Pointer Network \cite{Vinyals2015Pointer}. Instead of summing the attention of query-to-document, GA Reader \cite{Dhingra2017Gated} defined an element-wise product to endowing attention on each word of the document using the entire query representation to build query-specific representations of words in the document for accurate answer selection. \newcite{Wang2017Gated} employed gated self-matching networks (R-net) on passage against passage itself to refine passage representation with information from the whole passage. \newcite{Cui2016Attention} introduced an ``attended attention" mechanism (AoA) where query-to-document and document-to-query are mutually attentive and interactive to each other. 

\subsection{Augmented Word Embedding}
Distributed word representation plays a fundamental role in neural models \cite{Cai2016Neural,Qin2016A,Zhao2017B,Peters2018ELMO,He2018Syntax,Wang2018Graph,Bai2018deep,zhang2018NHD}. Recently, character embeddings are widely used to enrich word representations \cite{Kim2015Character,Yang2016Words,luong2016achieving,Huang2018Moon}. \newcite{Yang2016Words} explored a fine-grained gating mechanism (FG Reader) to dynamically combine word-level and character-level representations based on properties of the words. However, this method is computationally complex and it is not end-to-end, requiring extra labels such as NE and POS tags. \newcite{Seo2016Bidirectional} concatenated the character and word embedding to feed a two-layer Highway Network.

Not only for machine reading comprehension tasks, character embedding has also benefit other natural language process tasks, such as word segmentation \cite{Cai2017Fast}, machine translation \cite{luong2016achieving}, tagging \cite{yang2016multi,Li2018} and language modeling \cite{Verwimp2017Character,Miyamoto2016Gated}. However, character embedding only shows marginal improvement due to a lack internal semantics. Lexical, syntactic and morphological information are also considered to improve word representation \cite{Cao2016A,Bergmanis2017From}. \newcite{Bojanowski2016Enriching} proposed to learn representations for character $n$-gram vectors and represent words as the sum of the $n$-gram vectors. \newcite{Avraham2017The} built a model inspired by \cite{Joulin2016Bag}, who used morphological tags instead of $n$-grams. They jointly trained their morphological and semantic embeddings, implicitly assuming that morphological and semantic information should live in the same space. However, the linguistic knowledge resulting subwords, typically, morphological suffix, prefix or stem, may not be suitable for different kinds of languages and tasks. \newcite{Sennrich2015Neural} introduced the byte pair encoding (BPE) compression algorithm into neural machine translation for being capable of open-vocabulary translation by encoding rare and unknown words as subword units. Instead, we consider refining the word representations for both frequent and infrequent words from a computational perspective. Our proposed subword-augmented embedding approach is more general, which can be adopted to enhance the representation for each word by adaptively altering the segmentation granularity in multiple NLP tasks. 

\section{Conclusion}

This paper presents an effective neural architecture, called subword-augmented word embedding to enhance the model performance for the cloze-style reading comprehension task. The proposed SAW Reader uses subword embedding to enhance the word representation and limit the word frequency spectrum to train rare words efficiently. With the help of the short list, the model size will also be reduced together with training speedup. Unlike most existing works, which introduce either complex attentive architectures or many manual features, our model is much more simple yet effective. Giving state-of-the-art performance on multiple benchmarks, the proposed reader has been proved effective for learning joint representation at both word and subword level and alleviating OOV difficulties.

\bibliography{acl2018}
\bibliographystyle{acl}

\end{document}